\pdfoutput=1

\documentclass[11pt]{article}

\usepackage{acl}

\usepackage{graphicx} 
\usepackage{multirow}
\usepackage{amsmath}
\usepackage{threeparttable}
\usepackage{booktabs}

\usepackage{algorithm}
\usepackage{algorithmic}
\usepackage{amsfonts}
\usepackage{amssymb}

\usepackage{newfloat}
\usepackage{listings}
\usepackage{hyperref}
\usepackage{times}
\usepackage{latexsym}
\usepackage{makecell}
\usepackage{subfigure}

\usepackage[T1]{fontenc}

\usepackage[utf8]{inputenc}

\usepackage{microtype}

\usepackage{adjustbox}

\definecolor{CadmiumOrange}{rgb}{0.93,0.53, 0.18}
\definecolor{ForestGreen}{rgb}{0.13, 0.55, 0.13}

\title{VIRT: Improving Representation-based Text Matching via \\ Virtual Interaction}

\author{
  Dan Li \textsuperscript{\rm1}\thanks{\enspace Equal contribution.}, Yang Yang \textsuperscript{\rm2}\footnotemark[1], Hongyin Tang \textsuperscript{\rm2}, Jiahao Liu \textsuperscript{\rm2}, Qifan Wang \textsuperscript{\rm3},\\ \textbf{Jingang Wang \textsuperscript{\rm2}\thanks{\enspace Corresponding author.}, Tong Xu \textsuperscript{\rm1}\footnotemark[2], Wei Wu \textsuperscript{\rm2}, Enhong Chen \textsuperscript{\rm1}} \\
  \textsuperscript{\rm 1} University of Science and Technology of China
  \textsuperscript{\rm 2} Meituan 
  \textsuperscript{\rm 3} MetaAI \\
  \texttt{\{lidan528,tongxu,cheneh\}@mail.ustc.edu.cn} \\
  \texttt{\{yangyang113,tanghongyin,liujiahao12,wangjingang02\}@meituan.com} \\
  \texttt{wqfcr@fb.com,} \texttt{wuwei19850318@gmail.com} \\
}

\begin{document}
\sloppy\hyphenpenalty=5000
\maketitle

\begin{abstract}
Text matching is a fundamental research problem in natural language understanding.
Interaction-based approaches treat the text pair as a single sequence and encode it through cross encoders, while representation-based models encode the text pair independently with siamese or dual encoders.
Interaction-based models require dense computations and thus are impractical in real-world applications. Representation-based models have become the mainstream paradigm for efficient text matching.
However, these models suffer from severe performance degradation due to the lack of interactions between the pair of texts.
To remedy this, we propose a \textbf{V}irtual \textbf{I}nte\textbf{R}ac\textbf{T}ion mechanism (VIRT) for improving representation-based text matching while maintaining its efficiency.
In particular, we introduce an interactive knowledge distillation module that is only applied during training. It enables deep interaction between texts by effectively transferring knowledge from the interaction-based model. 
A light interaction strategy is designed to fully leverage the learned interactive knowledge.
Experimental results on six text matching benchmarks demonstrate the superior performance of our method over several state-of-the-art representation-based models. We further show that VIRT can be integrated into existing methods as plugins to lift their performances.
\end{abstract}
\section{Introduction}
Text matching aims to model the semantic correlation between a pair of texts, which is a fundamental problem in various natural language understanding applications.
For instance, in community question answering (CQA) \cite{cqa,cqasurvey} systems, a key component is to find similar questions from the database regarding a user question via question matching \cite{GuptaPEBMJS18,qqp}. 
Similarly, a dialogue agent \cite{welleck2019dialogue} needs to make logical inferences \cite{ConneauKSBB17,GaoYC21} between a user statement and some pre-defined hypotheses by predicting their entailment relations.





\begin{figure*}[t] 
\centering 
\includegraphics[width=0.95\textwidth]{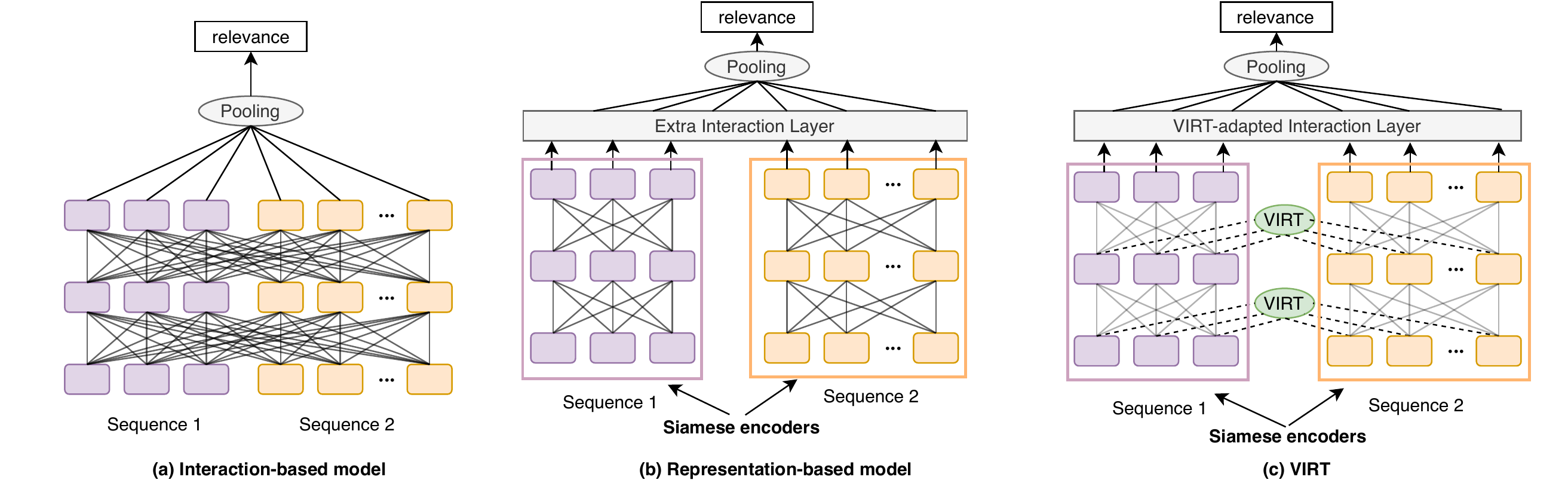} 
\caption{Schematic diagrams illustrating paradigms of text matching. The figure contrasts existing approaches (sub-figures (a) and (b)) with the proposed model (sub-figure (c)).
} 
\label{intro} 
\end{figure*}

Recently, the wide use of deep pre-trained Transformers \cite{vaswani2017attention} 
has made remarkable progress in text matching tasks \cite{RaffelSRLNMZLL20,Sentence-T5,DSI}.Two paradigms based on fine-tuned Transformer encoders are typically built:
interaction-based models and representation-based models, as illustrated in Figure~\ref{intro}(a) \& (b).
Interaction-based models (e.g., BERT~\citep{devlin-etal-2019-bert}) jointly encode the text pair, which allows the two text sequences to attend each other from the bottom layer to the top layer, resulting in effective matching signals. 
However, full interaction leads to high computational cost with large inference latency. In addition, text embedding can not be cached or pre-computed, which makes them impractical in many real-world scenarios. 
For example, in an E-commerce search system, it will cost dozens of days to score millions of query-product pairs with interaction-based models \cite{chen2020dipair}.
Representation-based models \citep{khattab2020colbert,Sentence-T5} encode two texts independently with siamese or dual encoders \cite{CerYKHLJCGYTSK18,reimers2019sentence},
which enable the offline-computing of text embeddings and thus significantly reduce the online latency. Unfortunately, independent encoding without any interaction fails to capture the correlation between the text pair, resulting in severe performance degradation.

To balance efficiency and efficacy, several works attempt to equip the siamese structure with late interaction modules. These late interactions are essentially light-weight interaction layers that fuse the two text embeddings from the individual encoders. A variety of late interaction strategies have been proposed, including MLP layers \cite{NEURIPS2021_4cc05b35}, cross-attention layers \cite{humeau2019poly} and Transformer layers \cite{cao2020deformer}, which obtain considerable improvements on different text matching tasks with reasonable costs.
However, these interaction modules are added after Siamese encoders, while interactions in the encoding process of Siamese encoders are still ignored, leaving a large performance gap compared to the interaction-based models.

In this work, 
we propose a Virtual InteRacTion (VIRT) mechanism with interactive knowledge distillation for improving representation-based text matching while keeping its efficiency.
Specifically,
Siamese encoders learn interactive information between the pair of texts by mimicking the full interaction, with transferred knowledge from the interaction-based models as guidance.
We employ the knowledge transfer as an attention map distillation during training, which is removed during inference to keep the Siamese property, and thus called ``virtual interaction''.
Moreover, we design a VIRT-adapted interaction strategy after Siamese encoding to further leverage the learnt interactive knowledge. Our proposed VIRT is illustrated in Figure~\ref{intro}(c).
Experimental results on six text matching benchmarks show the superior performance of VIRT over several state-of-the-art baselines. We
summarize the main contributions of this work as follows:
\begin{itemize}
    \item We propose a novel virtual interaction encoder for representation-based text matching, which effectively models the correlation between a pair of texts without additional inference cost. To the best of our knowledge, it is the first work that introduces interaction into the encoding process of Siamese encoders.
    \item We develop an interactive knowledge distillation module, which enables deep interaction by transferring knowledge from the interaction-based model. In addition, we design a VIRT-adapted interaction layer to further leverage the learnt interactive knowledge.
    \item Extensive experiments show that the proposed VIRT outperforms previous SOTA representation-based models, and maintains inference efficiency. The results also indicate that VIRT can be easily integrated into any representation-based text matching models for boosting their performance.
\end{itemize}

\section{Related Work}

\paragraph{Text Matching Models}
Text matching models typically take two textual sequences as input and determine their semantic relationship.
Early works perform keyword-based matching such as TF-IDF and BM25 \cite{J2009Integrating}. These methods rely on manually defined discrete features, thus usually fail to evaluate the semantic relevance of texts.
With the development of deep learning, a large variety of neural models have been proposed for text matching, which use recurrent neural networks \cite{2016Sequential,2017LearningM,2016aNMM} and convolutional neural networks \cite{hu2014convolutional} as the backbone, and encode textual sequences into semantic embeddings for fine-grained matches. 

Recently Transformer-based models \cite{BaoZHLMVDC19,LiZHWYL20} leverage self-attention to achieve promising performance on several text matching tasks \cite{DBLP:conf/acl/TangSJWZW20,DBLP:conf/naacl/QuDLLRZDWW21,DBLP:conf/iclr/XiongXLTLBAO21}. 
Generally, these models can be classified into interaction-based models \cite{LogeswaranL18,devlin-etal-2019-bert} and representation-based models \cite{reimers2019sentence}. 
As a typical interaction-based model, BERT \cite{devlin-etal-2019-bert} concatenates the text pair as the input and uses its \texttt{[CLS]} token embedding to predict the matching \cite{nogueira2019passage}. In contrast,
representation-based models utilize dual encoders to encode the pair of texts individually, which achieve high inference efficiency by pre-computing and storing all text embeddings in the database.
However, there is usually a large performance degradation compared to interaction-based models. More recently, late interactions with light attention layers \cite{humeau2019poly,khattab2020colbert,cao2020deformer} have been introduced after dual encoders to balance efficiency and efficacy. However, rich interactive information between the text pair is still ignored during encoding.
\paragraph{Knowledge Distillation}
Knowledge distillation \cite{hinton2015distilling,tang2019distilling} is to transfer knowledge from a teacher model with better quality to a less complex student model. 
Various works \cite{jiao2020tinybert,sanh2019distilbert,sun2019patient,sun2020mobilebert} have been proposed to compress BERT to a tiny structure with fewer Transformer layers and smaller hidden size through distilling predicted logits and hidden states. 
There are several recent distillation works that are closely related to our work.
DiPair \cite{chen2020dipair} performs extra interaction through a light Transformer layer, and distills predicted logits from the interaction-based model. 
Deformer \cite{cao2020deformer} adopts multiple Transformer-based interaction layers and distills the representations as well as the predicted logits from the interaction-based model.
However, these methods merely distill logits/representations from interaction-based models to the late interaction layer of representation-based models.
In contrast, VIRT distills the attention map from the interaction-based model directly to the encoding process of Siamese encoders, which transfers interactive knowledge more effectively.




\section{Methodology}


\begin{figure*}[ht] 
\centering 
\includegraphics[width=1.0\textwidth]{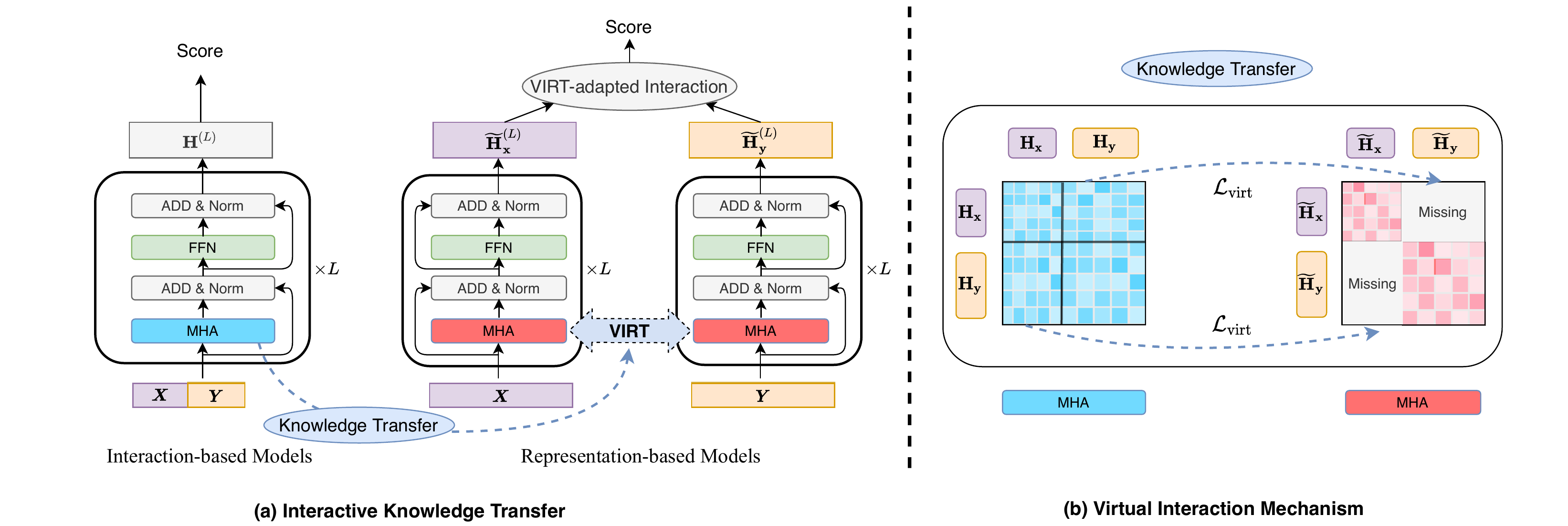} 
\caption{The proposed VIRT model architecture. (a) Interactive knowledge transfer procedure by distilling the attention map from the interaction-based model.
(b) VIRT mechanism details.} 
\label{model} 
\end{figure*}

\subsection{Preliminaries}
\paragraph{Interaction-based Models}

Given two textual sequences $\boldsymbol{X}=[\mathbf{x}_{1}; \ldots; \mathbf{x}_{m}]$ and $\boldsymbol{Y}=[\mathbf{y}_{1}; \ldots; \mathbf{y}_{n}]$ as input, the interaction-based models concatenate $\boldsymbol{X}$ and $\boldsymbol{Y}$ into $[\boldsymbol{X};\boldsymbol{Y}]$, and encode $[\boldsymbol{X};\boldsymbol{Y}]$ with a Transformer encoder \cite{devlin-etal-2019-bert}: $\mathbf{H}^{(L)}=\text{Enc}\left([\boldsymbol{X};\boldsymbol{Y}]\right)$. Each layer of Transformer consists of two residual sub-layers: a multi-head attention operation (MHA) (i.e., Eq. \ref{1a},  Eq. \ref{1b}) and a feed-forward network (FFN) (i.e., Eq. \ref{1c}):
\begin{subequations}
\begin{align}
&\mathbf{M}^{(l)} = \operatorname{softmax}\left( \mathrm{Att}( \mathbf{Q}^{(l)},  \mathbf{K}^{(l)}) \right) \label{1a}, \\
&\hat{\mathbf{H}}^{(l)}=\mathrm{LN}\left(\mathbf{M}^{(l)}_{}\mathbf{V}^{(l)} + \mathbf{H}^{(l-1)} \right) \label{1b}, \\
&\mathbf{H}^{(l)}=\mathrm{LN}\left(\mathrm{FFN}\left(\hat{\mathbf{H}}^{(l)}\right) + \hat{\mathbf{H}}^{(l-1)}\right). \label{1c}
\end{align}
\end{subequations}
where $\text {Att}(\mathbf{Q},\mathbf{K}) = \frac{\mathbf{Q}\mathbf{K}^T}{\sqrt{d}}$ is used to compute the attention map $\mathbf{M}$. $d$ is the dimension of hidden states. 
$\mathbf{H}^{(l)}$ is the intermediate representation from the $l$-th layer.
$\mathbf{Q} = \mathbf{HW_Q}$, $\mathbf{K}=\mathbf{HW_K}$ and $\mathbf{V}=\mathbf{HW_V}$ are the query, key and value matrices.
$\mathrm{LN}(\cdot)$ refers to the Layer-Normalization operation.
The interaction-based models are able to encode interactive information into the representations of $\boldsymbol{X}$ and $\boldsymbol{Y}$ through the full attention mechanism. 

\paragraph{Representation-based Models}
In contrast to interaction-based models, representation-based models encode $\boldsymbol{X}$ and $\boldsymbol{Y}$ individually through two independent Siamese Transformer encoders: $\widetilde{\mathbf{H}}^{L}_{\mathbf{x}} = \text{Enc}_{\mathbf{x}}\left(\boldsymbol{X}\right)$, and $\widetilde{\mathbf{H}}^{L}_{\mathbf{y}} = \text{Enc}_{\mathbf{y}}\left(\boldsymbol{Y}\right)$.
These models are very efficient, especially for downstream retrieval tasks: 1) they do not need to conduct pairwise encoding. 2) text embedding for the corpus can be pre-computed.
However, since there is no interaction between $\boldsymbol{X}$ and $\boldsymbol{Y}$ during encoding, fine-grained interactive information would be lost in representation-based models, resulting in significant performance degradation.

\subsection{VIRT}
The major weakness of representation-based models is lacking interaction when individually encoding two input sequences. 
Essentially, the interaction-based models perform interaction through the attention mechanism, and compute a unified attention map using both $\boldsymbol{X}$ and $\boldsymbol{Y}$. On the other hand, the representation-based models compute two disjoint attention maps from $\boldsymbol{X}$ and $\boldsymbol{Y}$ respectively. In the following sections, we first present the details of the difference between these two types of models in terms of the MHA operation. 
Next, we introduce the VIRT mechanism which improves the representation-based models without extra inference cost.

\paragraph{MHA Analysis}
The MHA operation in interaction-based models is illustrated by the blue attention map in Figure~\ref{model}(b).
Specifically, the input representations $\mathbf{H}$ of the $l$-th layer in interaction-based models could be decomposed to the $\boldsymbol{X}$-part and the $\boldsymbol{Y}$-part, i.e., $\mathbf{H} = [\mathbf{H}_{\mathbf{x}};\mathbf{H}_{\mathbf{y}}]$, where $\mathbf{H}_{\mathbf{x}}=[\mathbf{h}_1; ...; \mathbf{h}_{m}]$ and $\mathbf{H}_{\mathbf{y}}=[\mathbf{h}_{m+1}; ...; \mathbf{h}_{m+n}]$.
Note that we omit the superscript $l$ here for the simplicity of the presentation. 
In the attention map computation, the query and key matrices could also be rewritten as the combination of the $\boldsymbol{X}$-part and the $\boldsymbol{Y}$-part, i.e., $\mathbf{Q}=[\mathbf{Q}_{\mathbf{x}};\mathbf{Q}_{\mathbf{y}}]$ and $\mathbf{K}=[\mathbf{K}_{\mathbf{x}};\mathbf{K}_{\mathbf{y}}]$.
According to Eq.~\ref{1a}, the final attention score before the $\operatorname{softmax}(\cdot)$ operation (denoted as $\mathbf{S}$) could be decomposed as the following partitioned matrix:
\begin{equation}
    \begin{aligned}
        &\mathbf{S}= \mathrm {Att}\left( [\mathbf{Q}_{\mathbf{x}};  \mathbf{Q}_{\mathbf{y}}],  [\mathbf{K}_{\mathbf{x}};  \mathbf{K}_{\mathbf{y}}] \right) \\
        & = \left[\begin{array}{cc} 
        \mathrm {Att}(\mathbf{Q}_{\mathbf{x}}, \mathbf{K}_{\mathbf{x}})\ \  & \mathrm {Att}(\mathbf{Q}_{\mathbf{x}}, \mathbf{K}_{\mathbf{y}}) \\ 
        \mathrm {Att}(\mathbf{Q}_{\mathbf{y}}, \mathbf{K}_{\mathbf{x}})\ \  & \mathrm {Att}(\mathbf{Q}_{\mathbf{y}}, \mathbf{K}_{\mathbf{y}})
    \end{array}\right]\\
        & = \left[\begin{array}{cc} 
        \mathbf{S}_{\mathbf{x}\rightarrow\mathbf{x}}\ \  & \mathbf{S}_{\mathbf{x}\rightarrow\mathbf{y}} \\ 
        \mathbf{S}_{\mathbf{y}\rightarrow\mathbf{x}}\ \  & \mathbf{S}_{\mathbf{y}\rightarrow\mathbf{y}}
    \end{array}\right].
    \end{aligned}
\end{equation}
In particular, $\mathbf{S}_{\mathbf{x}\rightarrow\mathbf{x}}\in \mathbb{R}^{m \times m}$ and $\mathbf{S}_{\mathbf{y}\rightarrow\mathbf{y}}\in \mathbb{R}^{n \times n}$ are the MHA operations performed in $\boldsymbol{X}$ or $\boldsymbol{Y}$ only, which correspond to the MHA operations in representation-based models.
$\mathbf{S}_{\mathbf{x}\rightarrow\mathbf{y}}\in \mathbb{R}^{m \times n}$ and $\mathbf{S}_{\mathbf{y}\rightarrow\mathbf{x}}\in \mathbb{R}^{n \times m}$ represent the interactions between $\boldsymbol{X}$ and $\boldsymbol{Y}$ in interaction-based models, which are responsible for enriching the representations with interactive information. However, these interactions are missing in representation-based models, as illustrated by the missing attention maps in Figure~\ref{model}(b).

\paragraph{Interactive Knowledge Transfer}
In order to bring the missing interaction back and bridge the performance gap, we let representation-based models mimic the interactions as:
\begin{equation}
\label{xtoy}
\begin{aligned}
&\widetilde{\mathbf{M}}_{\mathbf{x}\rightarrow\mathbf{y}} = \operatorname{softmax}\left( \mathrm{Att}( \widetilde{\mathbf{Q}}_{\mathbf{x}}, \widetilde{\mathbf{K}}_{\mathbf{y}}) \right), \\
&\widetilde{\mathbf{M}}_{\mathbf{y}\rightarrow\mathbf{x}} = \operatorname{softmax}\left( \mathrm{Att}( \widetilde{\mathbf{Q}}_{\mathbf{y}}, \widetilde{\mathbf{K}}_{\mathbf{x}} )\right), \\
\end{aligned}
\end{equation}
 where $\widetilde{\mathbf{M}}_{\mathbf{x}\rightarrow\mathbf{y}}$ denotes the attention map which is generated by $\widetilde{\mathbf{H}}_{\mathbf{x}}$ attending to $\widetilde{\mathbf{H}}_{\mathbf{y}}$, and similar for $\widetilde{\mathbf{M}}_{\mathbf{y}\rightarrow\mathbf{x}}$. 
These two additional attention maps represent the missing interactive signals in representation-based models, which are responsible for updating the representations.
However, they cannot be directly calculated from the dual encoders in representation-based models, resulting in less effective text embeddings.

To close the performance gap between representation-based and interaction-based models, we propose to align the missing attention maps with their counterparts that have already existed in interaction-based models. Intuitively, the attention maps in the interaction-based models can guide the learning of the representations to evolve towards an interaction-rich direction as if the representations have interacted with each other during the encoding process. By this means, we distill the knowledge in interaction and transfer it into the dual encoders without any extra computational cost in inference. That is why we call the mechanism ``virtual interaction''.

Concretely, we employ a trained interaction-based model as the teacher and distill the knowledge to a representation-based student model. 
In each layer, we obtain the attention maps $\mathbf{M}_{\mathbf{x}\rightarrow\mathbf{y}}$ and $\mathbf{M}_{\mathbf{x}\rightarrow\mathbf{y}}$ from the interaction-based model and transfer these supervised interactive knowledge to guide the learning of the representation-based model.
Formally, the goal is to minimize the $L_2$ distance across all layers between $(\widetilde{\mathbf{M}}_{\mathbf{x}\rightarrow\mathbf{y}}, \widetilde{\mathbf{M}}_{\mathbf{y}\rightarrow\mathbf{x}})$ and $({\mathbf{M}}_{\mathbf{x}\rightarrow\mathbf{y}}, {\mathbf{M}}_{\mathbf{y}\rightarrow\mathbf{x}})$:
\begin{equation}
\label{loss}
    \begin{aligned}
    \mathcal{L}_{\text {virt }} &= \frac{1}{2 L}\sum_{l=1}^{L} \left(\frac{1}{m} \left\|\widetilde{\mathbf{M}}^{(l)}_{\mathbf{x}\rightarrow\mathbf{y}} - \mathbf{M}^{(l)}_{\mathbf{x}\rightarrow\mathbf{y}} \right\|_2 \right. \\
    &+ \left. \frac{1}{n} \left\|\widetilde{\mathbf{M}}^{(l)}_{\mathbf{y}\rightarrow\mathbf{x}} - \mathbf{M}^{(l)}_{\mathbf{y}\rightarrow\mathbf{x}} \right\|_2 \right).
    \end{aligned}
\end{equation}
Note that the above distillation is only applied in the training stage to learn better dual encoders. This preserves the Siamese property of representation-based models without extra inference cost.

\subsection{VIRT-Adapted Interaction}
Through VIRT, interactive knowledge could be incorporated deeply into each encoding layer of the representation-based models.
However, after Siamese encoding, the representations of the last layer, i.e., $\widetilde{\mathbf{H}}^{(L)}_{\mathbf{x}}$ and $\widetilde{\mathbf{H}}^{(L)}_{\mathbf{y}}$, still cannot see each other, and thus lack explicit interaction. To make full use of the learnt interactive knowledge, we further design a VIRT-adapted interaction strategy, which fuses $\widetilde{\mathbf{H}}^{(L)}_{\mathbf{x}}$ and $\widetilde{\mathbf{H}}^{(L)}_{\mathbf{y}}$ under the guidance of the attention map learnt by VIRT. 

Specifically, we perform VIRT-adapted interaction between the $\widetilde{\mathbf{H}}^{(L)}_{\mathbf{x}}$ and $\widetilde{\mathbf{H}}^{(L)}_{\mathbf{y}}$ following the process in Eq.\ref{xtoy}. The generated attention maps are formulated as follows:
\begin{equation}
    \label{fusion}
    \begin{aligned}
    &\hat{\mathbf{M}}^{(L)}_{\mathbf{x}\rightarrow\mathbf{y}} = \operatorname{softmax}\left( \mathrm{Att}( \widetilde{\mathbf{H}}^{(L)}_{\mathbf{x}},  \widetilde{\mathbf{H}}^{(L)}_{\mathbf{y}} )\right), \\
    &\hat{\mathbf{M}}^{(L)}_{\mathbf{y}\rightarrow\mathbf{x}} = \operatorname{softmax}\left( \mathrm{Att}( \widetilde{\mathbf{H}}^{(L)}_{\mathbf{y}},  \widetilde{\mathbf{H}}^{(L)}_{\mathbf{x}} )\right), \\
    &\mathbf{u} = \text{Pool}\left(\hat{\mathbf{M}}^{(L)}_{\mathbf{x}\rightarrow\mathbf{y}}  \widetilde{\mathbf{H}}^{(L)}_{\mathbf{y}}\right), \\
    &\mathbf{v} = \text{Pool}\left(\hat{\mathbf{M}}^{(L)}_{\mathbf{y}\rightarrow\mathbf{x}}  \widetilde{\mathbf{H}}^{(L)}_{\mathbf{x}}\right), \\
    \end{aligned}
\end{equation}
where $\text{Pool}(\cdot)$ denotes the mean pooling operation. Eq.~\ref{fusion} employs the same interaction strategy as VIRT, and further utilizes learnt attention maps to update representations explicitly. Finally, we utilize simple fusion to make predictions:
\begin{equation}
    \label{predict}
    \begin{aligned}
       \mathbf{r} &= (\mathbf{u}, \mathbf{v}, \mathbf{u}-\mathbf{v}, \text{max}(\mathbf{u}, \mathbf{v})), \\
       {y} &= \operatorname{softmax}\left(\text{MLP}(\text{MLP}(\mathbf{r})+\mathbf{r})\right),
    \end{aligned}
\end{equation}
where $(,)$ is the concatenate operation, and MLP denotes the Multi-Layer Perceptron. The overall training objective is minimizing the combination of the task-specific supervision loss $\mathcal{L}_{\text{task}}$ and the distillation loss $\mathcal{L}_{\text{virt}}$:
\begin{equation}
\label{alphaloss}
    \mathcal{L} = \mathcal{L}_{\text{task}} + \alpha \mathcal{L}_{\text{virt}},
\end{equation}
where $\alpha$ is a hyper-parameter to weight the influence of virtual interaction. 
It is noteworthy that VIRT is a general strategy, and can be used to enhance any representation-based matching models, as will be shown in experiments.

\section{Experiments}
\subsection{Datasets}
We conduct an extensive set of experiments on three types of datasets, including three sentence-sentence matching tasks (MNLI, QQP, RTE), one question answering task (BoolQ) and two real-world query-passage matching tasks (Q2P, Q2A). 

An overview of all the datasets is provided in Table \ref{statistic_table}. The detailed statistics and average text lengths are presented. Note that the average length of Chinese is based on characters, and English is based on words. 

\begin{table}[ht]
\begin{adjustbox}{width=0.9\columnwidth,center}
  \begin{tabular}{l|c|c|c}
    \toprule
    \multirow{2}{*}{Dataset} & \# of pairs & AvgLen  & AvgLen \\
    & (Train / Dev) & TextA & TextB \\
    \midrule
    MNLI & 392,702 / 9,815 & 19.6 & 10.0 \\
    RTE & 2,490 / 277 & 43.0 & 8.6 \\
    QQP & 327,464 / 40,430 & 10.9 & 11.2 \\
    BoolQ & 9,427 / 9,427 & 8.8 &  92.7 \\
    Q2P & 110,000 / 13,960 & 6.0 & 57.6 \\
    Q2A & 519,821 / 11,440 & 3.8 & 160.0 \\
    
  \bottomrule
  \end{tabular}
 \end{adjustbox}
 \caption{Datasets statistics. (For GLUE and  SuperGLUE, the results on development sets are reported since they do not distribute labels for test sets. For Q2P and Q2A datasets, we construct development sets, which is non-overlapping with the training sets.))}
\label{statistic_table}
\end{table}

\begin{table*}
\begin{adjustbox}{width=0.81\width,center}
\begin{tabular}{lcccccccc}
    \toprule
    Model & MNLI & RTE & QQP & BoolQ & Q2P & Q2A & \makecell[c]{Inference Latency (times)} \\
    \midrule
    BERT-Base & 84.1  & 66.0  & 90.6 & 74.1 & 91.0 &91.0& 332.6ms (1.0x) \\
    \midrule
    Siamese BERT \cite{devlin-etal-2019-bert} & 60.2 & 53.3 & 80.1 & 70.5 & 73.2 &80.6  & 47ms (7.1x)\\
    DeFormer \cite{cao2020deformer} & 71.1 & 55.0 & 88.5 & 70.9 & 84.0 & 84.1   & 118ms (2.8x)\\
    DiPair \cite{chen2020dipair}  & 71.3 & 55.1 & 88.6 & 71.3 & 80.3 & 87.4  & 49.1ms (6.8x)\\
    Poly-encoder \cite{humeau2019poly}   & 74.5 & 57.2 & 88.5 & 70.9 & 83.5 & 88.3  & 68.2ms (4.9x)\\
    Sentence-T5 \cite{Sentence-T5}  & 75.9 & 59.2 & 90.3 & 72.0 & 85.7 &  81.9  & 47.5ms (7.0x)\\
    VIRT (ours) & \bf78.6 & \bf60.5 & \bf90.4 & \bf73.1 & \bf89.2 & \bf90.1& 66.5ms (5.0x)\\
    
    \bottomrule
\end{tabular}
\end{adjustbox}
\caption{Performance comparison on six datasets. 
Note that we only report online parts of inference latency, since the representation-based embeddings could be computed offline and  online latency in real-world scenarios is more concerning. Since models on these six datasets take a similar input setup, we report inference 
latency on BoolQ and omit the other five. Results are
statistically significant with p-value < 0.001.}
\label{perform_table}
\end{table*}


\vspace*{1\baselineskip}
\noindent{\bf MNLI}
\cite{williams2017broad} is a large-scale entailment classification dataset. The objective is to predict the relationship between a pair of sentences as entailment, neutral, or contradiction.

\noindent{\bf RTE}
\cite{rte} dataset comes from a series of annual competitions on textual entailment. The objective is to predict whether a given hypothesis is entailed by a given premise.

\noindent{\bf QQP}
\cite{qqp} is a large-scale sentence similarity dataset with question pairs from Quora. The task is to determine if the two questions have the same meaning.

\noindent{\bf BoolQ}
\cite{clark-etal-2019-boolq} is a question answering dataset for yes/no questions given question and document pairs.

\noindent{\bf Q2P}
is a binary classification task derived from the MSMARCO Passage Ranking data \cite{DBLP:conf/nips/NguyenRSGTMD16} containing $110$K query passage pairs. Given a (query, passage) pair, the goal is to predict whether the passage contains the answer for the query. The original dataset does not contain labeled negative samples. For each query, we sample the negative passage from the top-$100$ passages retrieved by BM25. 

\noindent{\bf Q2A}
is our internal dataset containing a huge amount of query-advertisement pairs. 
All the data are crawled from a Chinese E-commerce website and manually annotated. 
Given a (query, advertisement) pair, the goal is to predict the relevance between the advertisement and the query.

\subsection{Baselines}
We adopt several state-of-the-art representation-based matching models as our baselines.



\vspace*{1\baselineskip}
\noindent{\bf Siamese BERT}
\cite{devlin-etal-2019-bert} is a Siamese architecture that uses pre-trained BERT to separately produce embeddings of two inputs. The pooled output embeddings of two sequences are concatenated to give final predictions.


\noindent{\bf DeFormer}
\cite{cao2020deformer} is a decomposed BERT-based model, which splits the full self-attention into two independent self-attention in the lower layers of BERT while the upper layers are kept origin with full self-attention.

\noindent{\bf DiPair}
\cite{chen2020dipair} is a fast and distilled representation-based model for text matching. It performs extra interaction through a light transformer layer, which feeds with truncated embeddings output from the last encoder layer. 

\noindent{\bf Poly-encoders}
\cite{humeau2019poly} is a representation-based model for pairwise text matching which utilizes an attention mechanism to perform extra interaction after Siamese encoders. 

\noindent{\bf Sentence-T5}
\cite{Sentence-T5} learns sentence embeddings from text-to-text Transformers T5 \cite{T5}. The output embeddings of two sequences and their difference are concatenated to give final predictions.

\subsection{Experimental Setup}
\paragraph{VIRT setup}
We use BERT-base \cite{devlin-etal-2019-bert} as the encoder backbone of VIRT. The parameters are initialized with the pre-trained BERT-base model (uncased). We share all parameters between $\text{Enc}_{\mathbf{x}}(\cdot)$ and $\text{Enc}_{\mathbf{y}}(\cdot)$. We also take BERT-base as the interaction-based model, which is finetuned first, and used as the teacher model to transfer interaction knowledge to 
representation-based models. The pooling strategy of BERT-base at the prediction layer is fixed to mean pooling (instead of \texttt{[CLS]}), as we observe better performance on both BERT-base and all VIRT-enhanced representation-based models.

\paragraph{Implementation Details}
All baselines are initialized with pre-trained BERT-base parameters, and fine-tuned to achieve the best results on the validation sets. 
It is worth noting that we fix the total number of transformer layers for all models at $12$ to make a fair comparison, though some of the baselines such as DiPair~\cite{chen2020dipair} take fewer layers for extreme efficiency at the cost of performance. 
The first 8 and first 16 output token embeddings of $\boldsymbol{X}$ and $\boldsymbol{Y}$ are picked out as DiPair's input, which is the best setting reported from its paper.
The number of context vectors in Poly-encoders is 360.
For MNLI and QQP, we use the standard partition and metrics on the GLUE benchmark\footnote{https://gluebenchmark.com/}. For RTE and BoolQ, we follow the SuperGLUE\footnote{https://super.gluebenchmark.com/}. For Q2P and Q2A, we construct the dataset from MSMARCO Passage Ranking data and real-world E-commerce data using AUC-ROC as the evaluation metric.
We split $10\%$ of the training set for tuning hyper-parameters in these tasks, and report results on the original development split.

We implement all models with Tensorflow $1.15$ on Tesla V100 GPU (32GB memory). We set $\alpha$ as $1$ and the batch size as $28$. Training epochs for six tasks are set to $5$, $30$, $5$, $30$, $5$, $5$ respectively. Sequence length of two texts for six tasks are set to $(128, 128)$, $(64, 328)$, $(128, 128)$, $(64, 328)$, $(200, 200)$, $(16, 256)$ respectively. The learning rate is set to $5e-5$, with the warm-up ratio set to $0.1$. All models are optimized by Adam optimizer with $\beta_1 = 0.9$, $\beta_2 = 0.999$, $\epsilon=1e-8$. For measuring the online inference latency, we run the inference with the batch size set to $28$. We repeat each experiment $10$ times and report the metrics based on the average over these runs.
\begin{figure}
\centering
\includegraphics[width=0.9\linewidth]{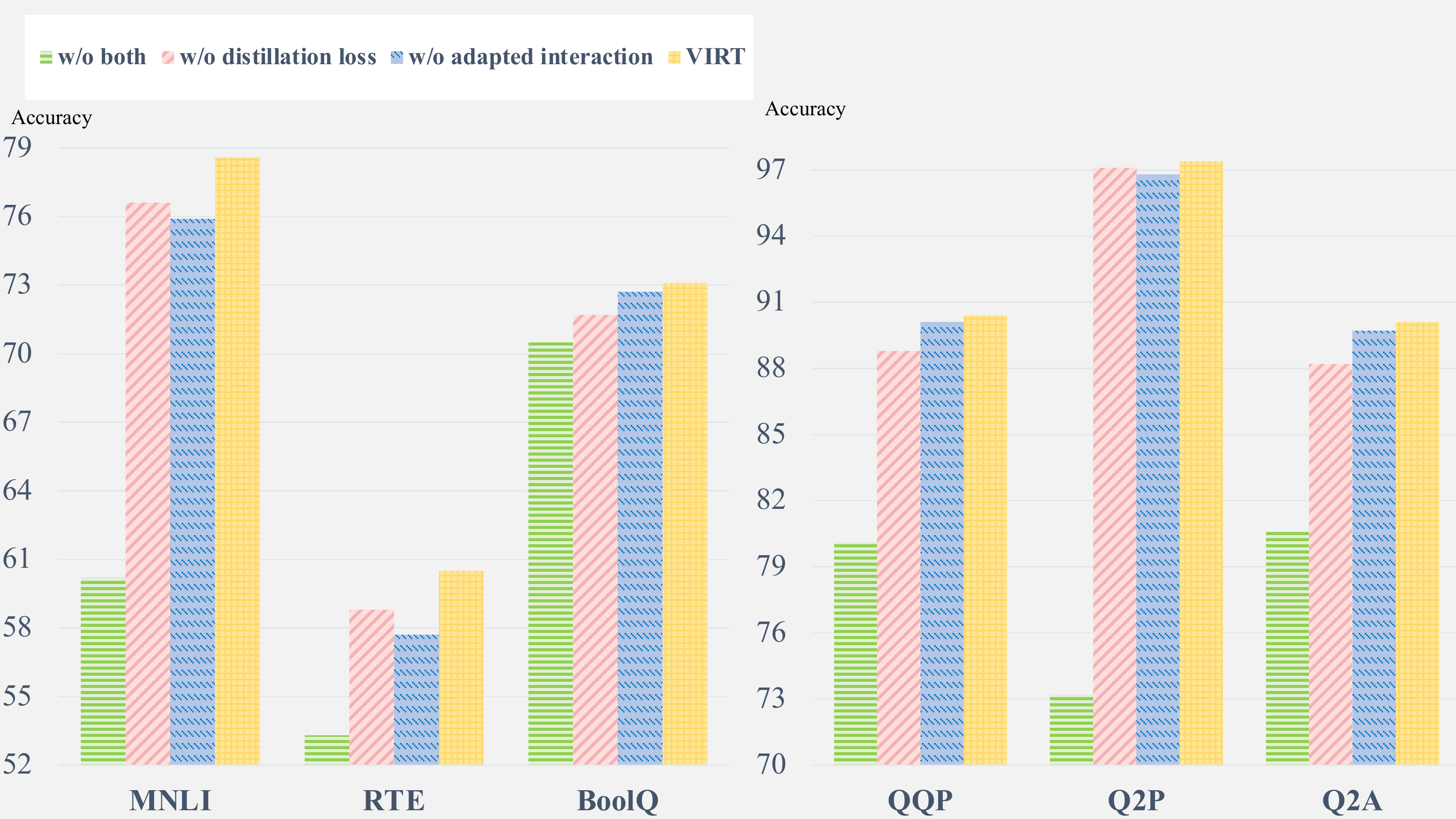}
\caption{Ablation analysis for different components on all datasets.} 
\label{ablation_study} 
\end{figure}

\subsection{Main Results}
The performance comparison of different methods is presented in Table \ref{perform_table}. 
BERT-base shows its effectiveness as a powerful interaction-based model.
Siamese BERT has a significant performance decline compared with BERT. 
DeFormer, DiPair, Poly-encoder and Sentence-T5 achieve considerable improvement compared with Siamese BERT.
Finally, VIRT achieves the best performance, outperforming all the representation-based baselines. It even obtains competitive results compared with the interaction-based BERT model. These results validate that VIRT is able to approximate the deep interaction modeling ability of the interaction-based models.

We further compare the inference latency on the BoolQ dataset across different models, which is also listed in Table \ref{perform_table}.  
According to the result, all representation-based models show significant speedup compared with the interaction-based models. 
The speedup mainly benefits from the Siamese encoder, which enables embeddings computed offline. 
Siamese BERT achieves the fastest inference speed, yet suffers from a severe performance decline. DeFormer gets relatively higher latency, due to the computation complexity of the extra interaction layers. 
Dipair truncates the sequence to a shorter length before the interaction layer, which produces an excellent speed-up in terms of online latency.
Poly-encoder and Sentence-T5 considerably improve the performance, at the cost of slightly increased computations.
Compared with all the baselines, our model shows superiority in terms of performance while keeping the high efficiency at the same time. 
Note that the inference latency is computed based on the average of all example pairs in an online manner. However, representation-based methods are able to pre-compute the embeddings of the corpus offline, and therefore dramatically reduce the inference time for downstream applications.

\subsection{Analysis and Discussion}
\subsubsection{Ablation Study}
To understand the impact of different components in VIRT, we conduct an ablation study by removing each component and retrain the models. 
In particular, ``w/o distillation loss'' means removing the optimization goal of Eq. \ref{loss}. ``w/o adapted interaction'' means removing the adapted interaction in Eq. \ref{fusion}, and using simple fusion for representation at the last layer as Eq. \ref{predict}. ``w/o both'' means remove both strategies simultaneously.
The results are shown in Figure \ref{ablation_study}.
The drop in performance without distillation or adapted interaction indicates the effectiveness of these two architectures. For MNLI and RTE, the performance drop caused by removing adapted interaction is more severe. Our hypothesis is that MNLI and RTE are natural language inference tasks, which require more fine-grained matching signals and rely heavily on explicit interaction. For QQP, BoolQ and Q2A, adapted interaction has less effect. However, distillation still brings substantial improvement, which further validates the effectiveness of incorporating interaction.

\begin{figure}[t] 
\centering
\includegraphics[width=0.9\linewidth]{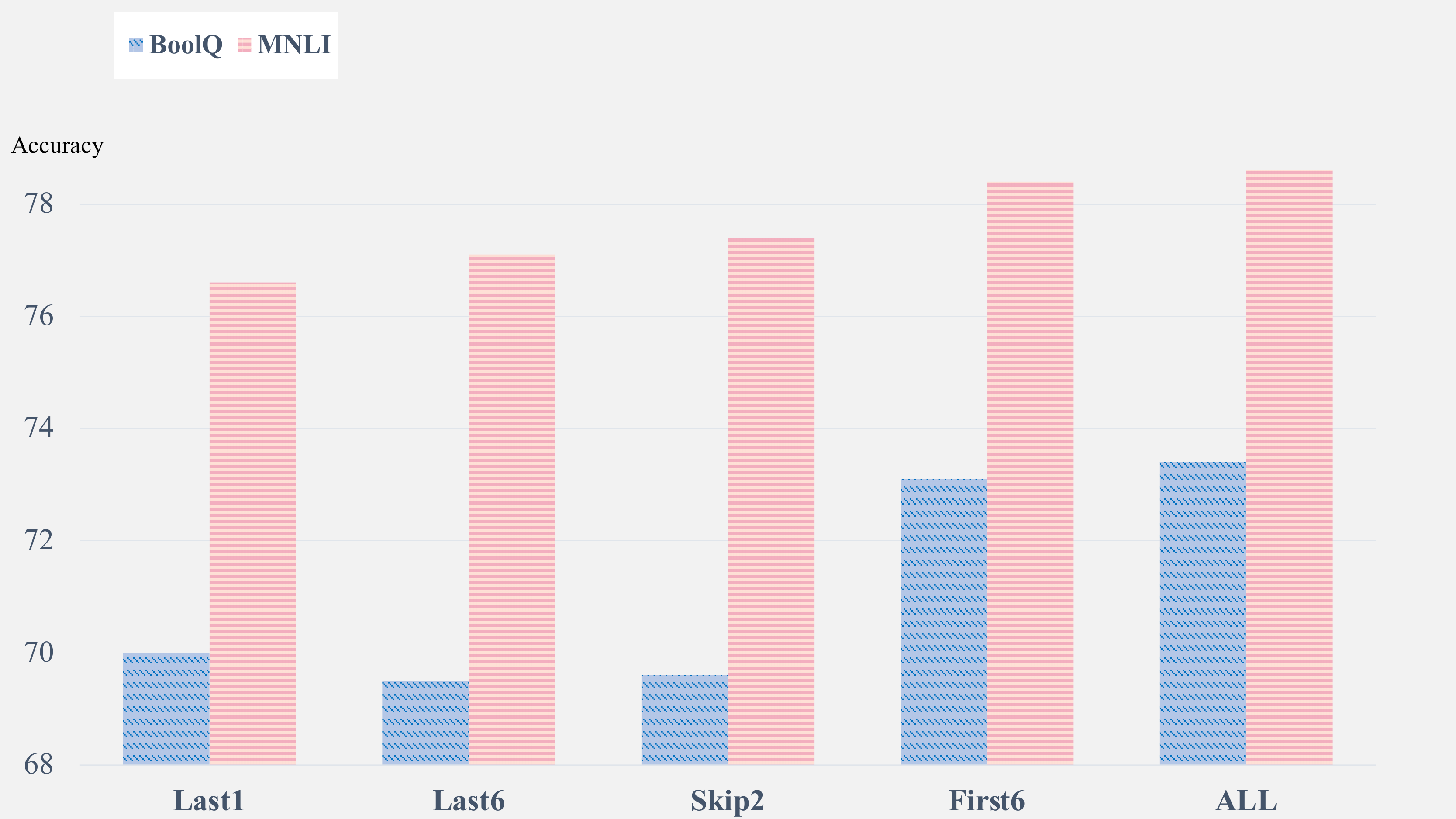}
\caption{Ablation study of applying VIRT to different encoder layers on MNLI and BoolQ.} 
\label{virtlayers} 
\end{figure}
\begin{table*}[t]
\begin{adjustbox}{width=0.8\width,center}
\begin{tabular}{lcccccccc}
    \toprule
    Model  & MNLI & RTE & QQP & BoolQ & Q2P & Q2A\\
    \midrule
    DeFormer + VIRT distillation& 72.3 (\textcolor{blue}{$\uparrow$1.2}) & 55.8 (\textcolor{blue}{$\uparrow$0.8})& 89.2 (\textcolor{blue}{$\uparrow$0.7}) & 71.8 (\textcolor{blue}{$\uparrow$0.9}) & 85.0 (\textcolor{blue}{$\uparrow$1.0}) & 86.1 (\textcolor{blue}{$\uparrow$2.0}) \\
    DiPair + VIRT distillation& 71.6 (\textcolor{blue}{$\uparrow$0.3}) & 55.3 (\textcolor{blue}{$\uparrow$0.2}) & 88.6 (\textcolor{blue}{-0.0}) & 71.8 (\textcolor{blue}{$\uparrow$0.5}) & 82.3 (\textcolor{blue}{$\uparrow$2.0}) &87.9 (\textcolor{blue}{$\uparrow$0.5})\\
    Poly-encoder + VIRT distillation& 75.3 (\textcolor{blue}{$\uparrow$0.8}) & 57.9 (\textcolor{blue}{$\uparrow$0.7}) & 89.2 (\textcolor{blue}{$\uparrow$0.7}) & 71.6 (\textcolor{blue}{$\uparrow$0.7})& 84.1 (\textcolor{blue}{$\uparrow$0.6}) &89.4 (\textcolor{blue}{$\uparrow$1.1}) \\
    Sentence-T5 + VIRT distillation& 77.2 (\textcolor{blue}{$\uparrow$2.3}) & 61.7 (\textcolor{blue}{$\uparrow$2.5}) & 90.8 (\textcolor{blue}{$\uparrow$0.5}) & 72.5 (\textcolor{blue}{$\uparrow$0.5}) &
    88.5 (\textcolor{blue}{$\uparrow$2.8}) & 83.5(\textcolor{blue}{$\uparrow$1.6})\\
    \bottomrule
\end{tabular}
\end{adjustbox}
\caption{Performance gain of applying VIRT distillation to different representation-based models. 
$\uparrow$ represents the performance gain.}
\label{virt_baseline}
\end{table*}

\subsubsection{Layer Importance}
In this set of experiments, we apply VIRT to different selected layers in the dual encoder to understand the importance of the interaction knowledge in different encoder layers. 
(1) VIRT-Last: only applying VIRT to the last $k$ layers.
(2) VIRT-First: only applying VIRT to the first $k$ layers.
(3) VIRT-Skip: applying VIRT to 1-in-$k$ layers.  
(4) VIRT-All: applying VIRT to all layers.

The results on MNLI and BoolQ are shown in Figure \ref{virtlayers}. 
It is not surprising to see that VIRT-All achieves the best performance over all the compared settings, showing the importance of the interaction for all layers. We observe that VIRT-First performs better than VIRT-Last and VIRT-Skip when all activating 6 layers, which indicates that interaction knowledge from the bottom layers plays a crucial role.
We also applied VIRT at the last one layer, referring to \cite{minilm} who claims distilling the last layer is enough. However, we find that when the teacher model and the student model are heterogeneous, merely distilling the information of the last one layer faces great performance degradation.

\subsubsection{Impact of VIRT Distillation}
To verify the generality and effectiveness of the proposed VIRT distillation, we further import it into the aforementioned representation-based models by applying the knowledge distillation to different baselines. The results are reported in Table \ref{virt_baseline}. According to the results, we can observe that 
VIRT distillation could be easily integrated into other representation-based text matching models to lift their performances.
Note that the results in Table \ref{virt_baseline} are different from the results of w/o adapted interaction in the ablation study. In the ablation study, we always leverage the fusion layer from Eq. \ref{predict}, which yields much better performances. Similar observations have been found in Sentence-T5 \cite{Sentence-T5}.
\begin{table}[h]
\begin{adjustbox}{width=0.85\width,center}
 \begin{tabular}{l|c}
    \toprule
    Model & MNLI  \\
    \midrule
    VIRT-BERT-Tiny$_2$ & 68.1 (\textcolor{blue}{$\uparrow$10.3)} \\
    VIRT-BERT-Mini$_4$ & 70.9 (\textcolor{blue}{$\uparrow$11.8)} \\
    VIRT-BERT-Small$_4$ & 73.6 (\textcolor{blue}{$\uparrow$13.5}) \\
    VIRT-BERT-Medium$_8$ &  74.5 (\textcolor{blue}{$\uparrow$14.3}) \\
    VIRT-BERT-Large$_{24}$ &  79.3 (\textcolor{blue}{$\uparrow$15.4})\\
  \bottomrule
  \end{tabular}
\end{adjustbox}
\caption{Performance gain of applying VIRT distillation to models with different configurations.}
\label{sizesmodels}
\end{table}
\subsubsection{Different Model Configurations}
We apply VIRT (including VIRT distillation and VIRT-adapted interaction) to pre-trained models with different sizes to show its robustness on different numbers of encoder layers. We conduct experiments using BERT-Tiny($2/128$), BERT-Mini($4/256$), BERT-Small($4/512$), BERT-Medium($8/512$), BERT-Base($12/768$) and BERT-Large($24/1024$) on the MNLI dataset, where $a/b$ means the number of encoder layers is $a$ and the dimension of hidden representation is $b$. The results are reported in Table~\ref{sizesmodels}. It can be seen from the results that VIRT yields better performance on all size of the pre-trained models, which is consistent with the observations from the main results.



\subsubsection{Impact of $\alpha$}
\begin{figure}[ht] 
\centering
\includegraphics[width=0.9\linewidth]{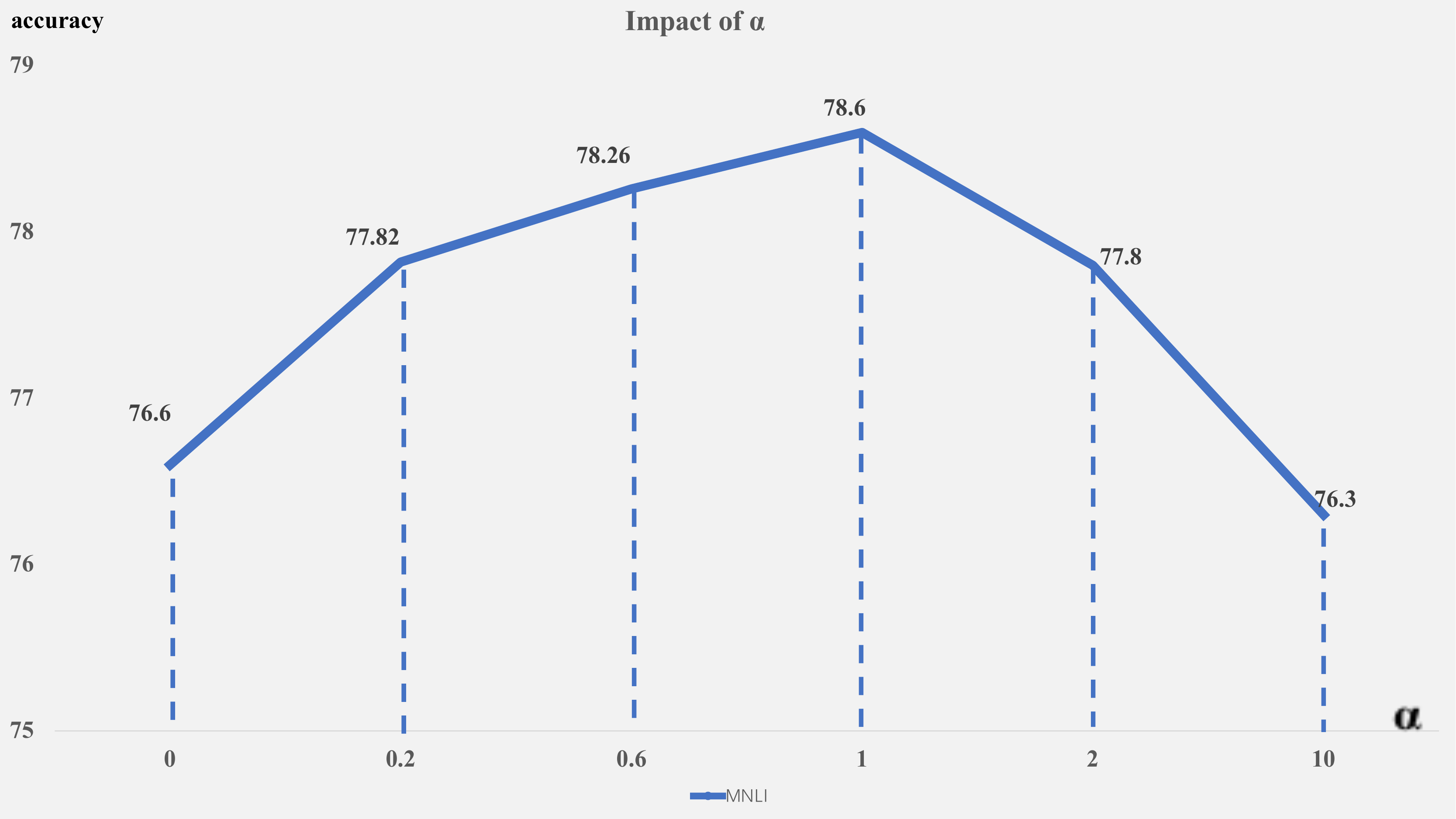}
\caption{Impact of $\alpha$ on MNLI.} 
\label{alphaexp}
\end{figure}

For our proposed VIRT approach, we conduct additional parameter search over $\alpha$ from $\{0, 0.2, 0.6, 1, 2, 10\}$  in Eq.~\ref{alphaloss} on the MNLI task.  The experimental results are shown in Figure~\ref{alphaexp}. From the results, it is clear that VIRT with $\alpha = 1$ achieves the best performance among all the $\alpha$ values, which illustrated that the $\mathcal{L}_{\text{virt}}$ is as important as $\mathcal{L}_{\text{task}}$. We also observe that the performance of VIRT is relatively stable with a wide range of $\alpha$, e.g., from 0.6 to 1.

\begin{figure}[ht] 
\centering 
\subfigure[The attention matrix of interaction-based model]{
\includegraphics[width=0.5\textwidth]{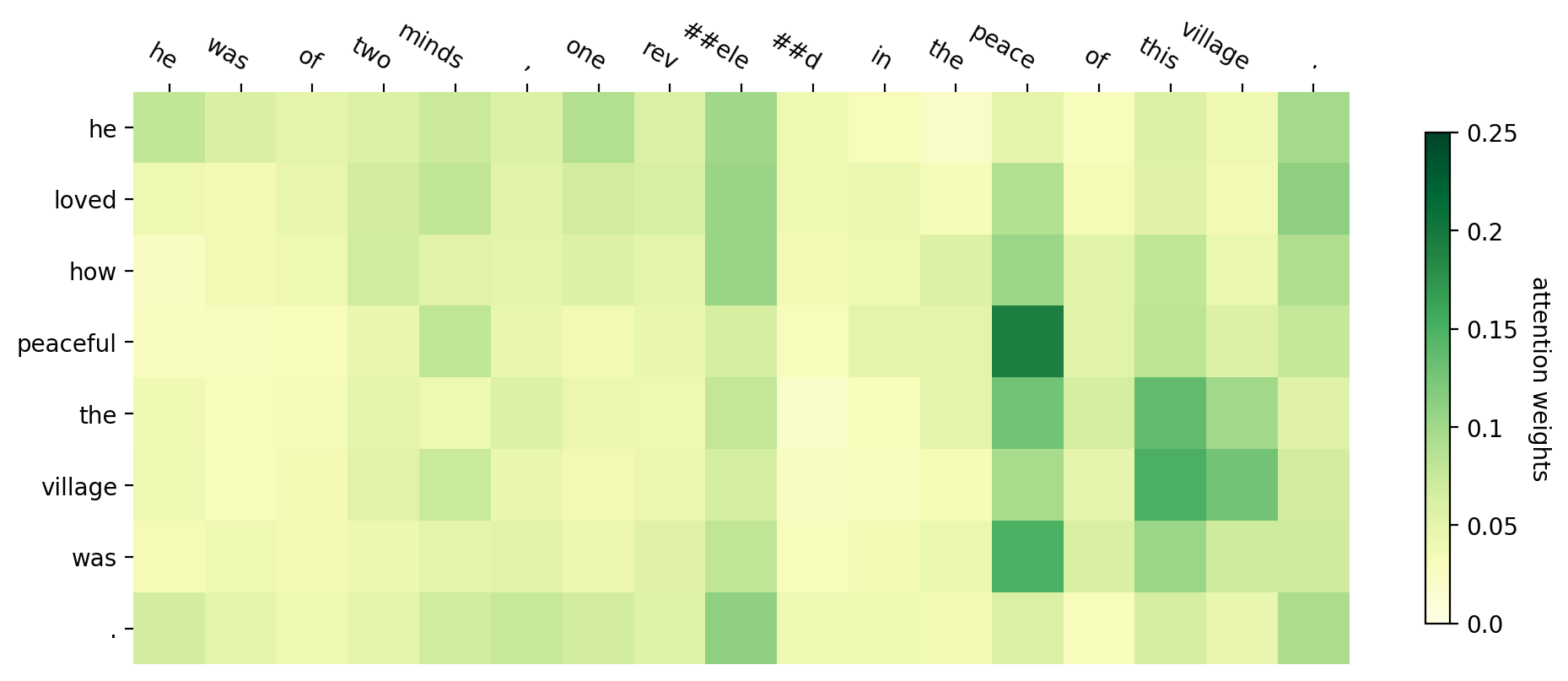}
\label{case1}
}
\subfigure[The attention matrix of representation-based model with VIRT distillation.]{
\includegraphics[width=0.5\textwidth]{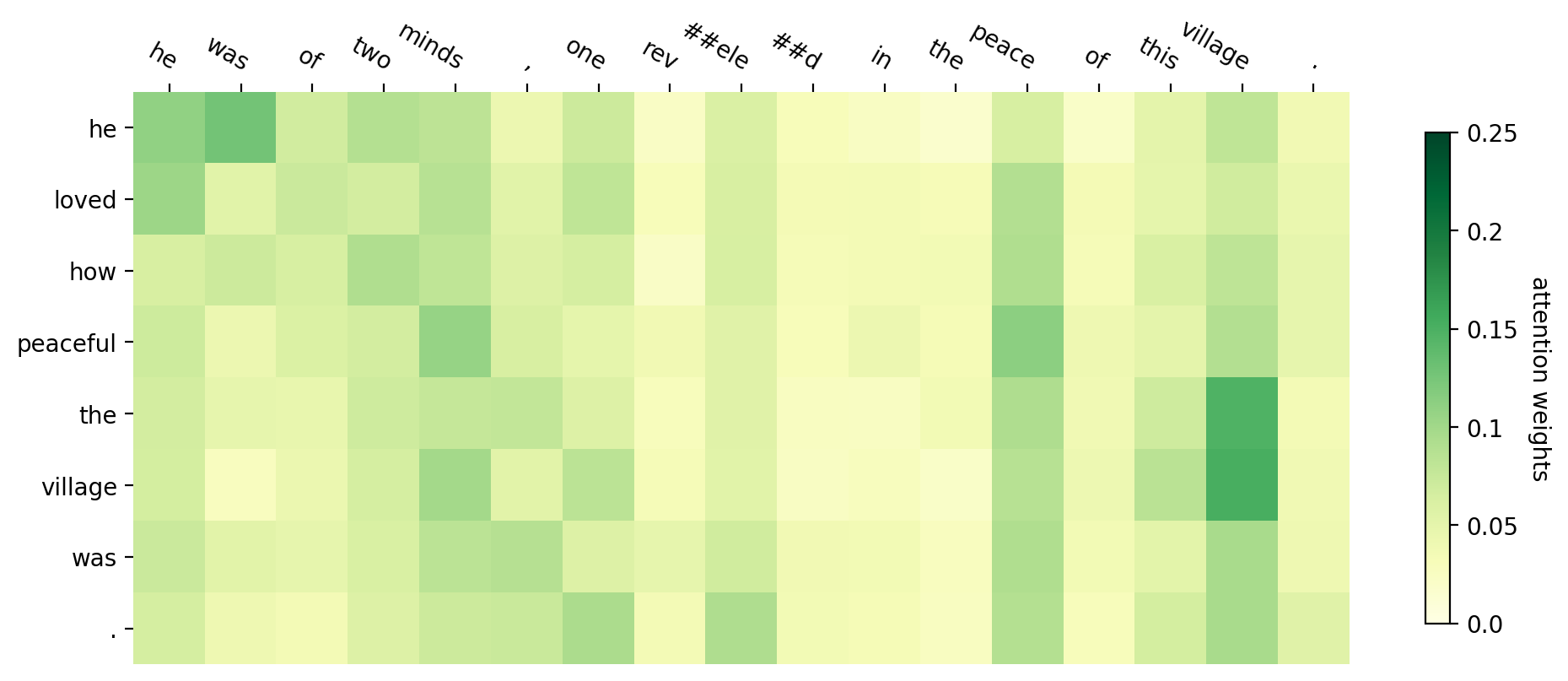} 
\label{case2}
}
\subfigure[The attention matrix of representation-based model without VIRT distillation.]{
\includegraphics[width=0.5\textwidth]{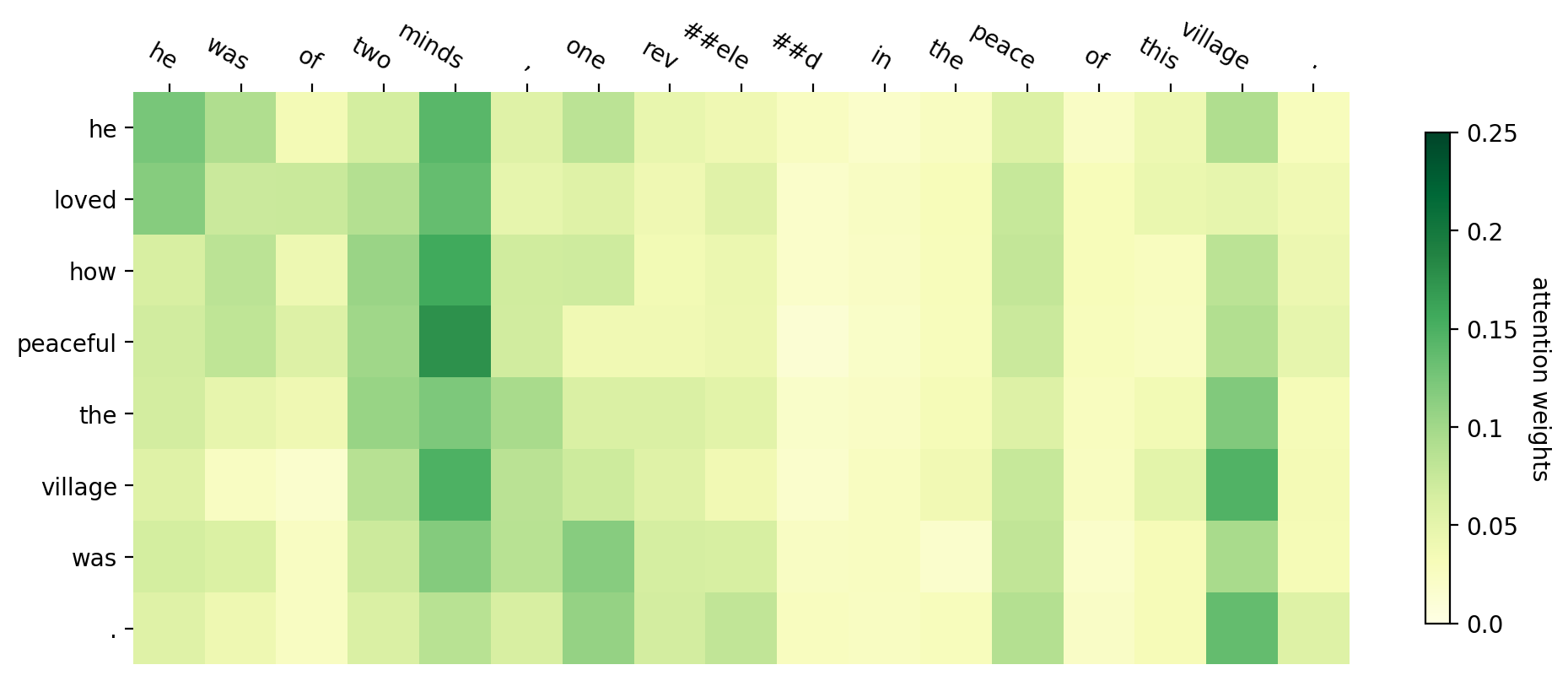}
\label{case3}
}
\caption{Visualization of the attention matrices.}
\end{figure}

\subsection{Case Study}
To show the effect of VIRT distillation in a more intuitive way, we visualize the attention matrices of different models. Specifically, we choose an example from the MNLI dataset and plot the corresponding attention matrices of the interaction-based model and the representation-based model with/without VIRT distillation. As shown in Figure \ref{case1}-\ref{case3}, the attention matrix with VIRT distillation is more consistent to the interaction-based model than the model without VIRT. In particular, the interaction-based model aligns ``peaceful'' with ``peace'' which can be learnt by VIRT whereas the representation-based model misses this information. As a result, the representation-based model without VIRT fails to predict the two sentences as ``neutral'' relationship.

\section{Conclusion}


Representation-based models are widely used in text matching tasks due to their high efficiency while under-performing the interaction-based ones caused by lacking interaction.
Previous works often introduce extra interaction layers while the interaction in Siamese encoders is still missing.
In this paper, we propose a virtual interaction (VIRT) mechanism , which could approximate the interactive modeling ability by distilling the attention map from interaction-based models to the Siamese encoders of representation-based models, with no additional inference cost. The proposed VIRT, which employs knowledge distillation as well as adapted interaction strategy, achieves state-of-the-art performance among existing representation-based models on several text matching tasks. 

\section*{Limitations}
Although the proposed VIRT mechanism enhances the performance of dual encoder architectures and achieves new SOTA on several datasets, two limitations are presented and discussed in this section.
First, in comparison to the vanilla dual encoder models such as Sentence-BERT, the training cost of VIRT is higher due to its introduction of virtual interaction distillation computation (i.e., the computational cost of distillation loss).
Second, the performance of VIRT is highly correlated with the performance of the interaction-based teacher.
Stronger teacher usually leads to the dual encoder student with higher performance.

\section*{Acknowledgements}
This work was supported by the grants from National Natural Science Foundation of China (No.U20A20229, 62072423), CAAI-Huawei MindSpore Open Fund (CAAIXSJLJJ-2021-007B), and the USTC Research Funds of the Double First-Class Initiative (No.YD2150002009).

\bibliography{Content/ref.bib}
\bibliographystyle{acl_natbib}

\clearpage

\appendix

\clearpage

\end{document}